\def\year{2020}\relax
\documentclass[letterpaper]{article} 
\usepackage{aaai20}  
\usepackage{times}  
\usepackage{helvet} 
\usepackage{courier}  
\usepackage[hyphens]{url}  
\usepackage{graphicx} 
\usepackage{graphicx}  
\usepackage{multirow}
\usepackage{xspace}
\usepackage{adjustbox}
\usepackage{enumitem}
\title{REST: A Thread Embedding  Approach for Identifying and Classifying User-specified Information in Security Forums}
\author{Joobin Gharibshah  ,
Evangelos E. Papalexakis , Michalis Faloutsos\\
University of California - Riverside, CA\\
900 University Ave, Riverside, California 92557\\
jghar002,epapalex,michalis@cs.ucr.edu}

\newcommand{\hide}[1]{}

\newcommand{\OffComm}{OffensiveCommunity\xspace}
\newcommand{\OffCommShort}{OffensComm.\xspace}
\newcommand{\HTS}{HackThisSite\xspace}

\newcommand{\Ethical}{EthicalHackers\xspace}
\newcommand{\EthicalShort}{EthicHack\xspace}

\newcommand{\TT}{Hacks\xspace}
\newcommand{\PS}{Services\xspace}
\newcommand{\AN}{Alerts\xspace}
\newcommand{\AG}{Experiences\xspace}

\newcommand{\myalg}{REST\xspace}

\newcommand{\ThSim}{T_{sim}\xspace}  
\newcommand{\ThWord}{T_{key}\xspace}  
\newcommand{\mdim}{m\xspace}  
\newcommand{\pro}{P\xspace}  
\newcommand{\tidx}{r\xspace}  
\newcommand{\classw}{w\xspace}  
\newcommand{\classi}{k\xspace}  

\newcommand{\word}{v\xspace}  

\newcommand{\WordClass}{WordClass\xspace}

\newcommand{\Words}{W\xspace}  
\newcommand{\Dict}{D\xspace}  

\newcommand{\WS}{WS\xspace}  

\begin{document}
%



\maketitle
\begin{abstract}
How can we extract useful information from a security forum?
We focus on identifying threads of interest to a security professional: (a) alerts of worrisome events, such as attacks, (b) offering of malicious services and products, (c) hacking information to perform malicious acts, and (d) useful security-related experiences.
The analysis of security forums is in its infancy despite several promising recent works.
Novel approaches are needed to address the challenges in this domain: (a) the difficulty in specifying the ``topics" of interest efficiently,  and (b) the unstructured and informal nature of the text.
We propose, \myalg, a systematic methodology to: (a) identify threads of interest based on a, possibly incomplete, bag of words, and (b) classify them into one of the four classes above.
The key novelty of the work is a multi-step weighted embedding approach: we project words, threads and classes in appropriate embedding spaces and establish relevance and similarity there. 
We evaluate our method with real data from three security forums with a total of 164k posts and 21K threads.
 First, \myalg robustness to initial keyword selection
 can extend the user-provided keyword set and thus, it can recover from missing keywords. Second, \myalg categorizes the threads into the classes of interest with superior accuracy compared to five other methods: \myalg exhibits  
  an accuracy between 63.3-76.9\%.
 We see our approach as a first step for harnessing the wealth of information of online forums in a user-friendly way, since the user can loosely specify her keywords of interest.

\end{abstract}


\section{Introduction}

Security forums hide a wealth of information, but mining it requires
novel methods and tools.
The problem is driven by practical forces: there is useful information that could help improve security, but the volume of the data requires an automated method.
The challenge is that there is a lot of ``noise",
there is lack of structure, and an abundance of informal and hastily written text. At the same time,  security analysts need receive
focused and categorized information, which can help their task of
shifting through it further. We define the problem more specifically below.

Given a security forum, we want to extract threads of interest to a security analyst. 
We consider two associated problems that together provide a complete solution.
First, the input is all the data of a forum, and the user specifies its interest
by providing one or more bag-of-words of interest. Arguably, providing keywords is a relatively easy task for the user. 
The goal is to return all the threads that are of interest to the user,
and we use the term {\bf relevant} to indicate such threads.
A key challenge here is how to create a robust solution that
is not overly sensitive to the omission of potentially important keywords. We use the term {\bf identification} to refer to this problem.

\begin{figure*}[t!]
  \includegraphics[keepaspectratio, width=1\textwidth]{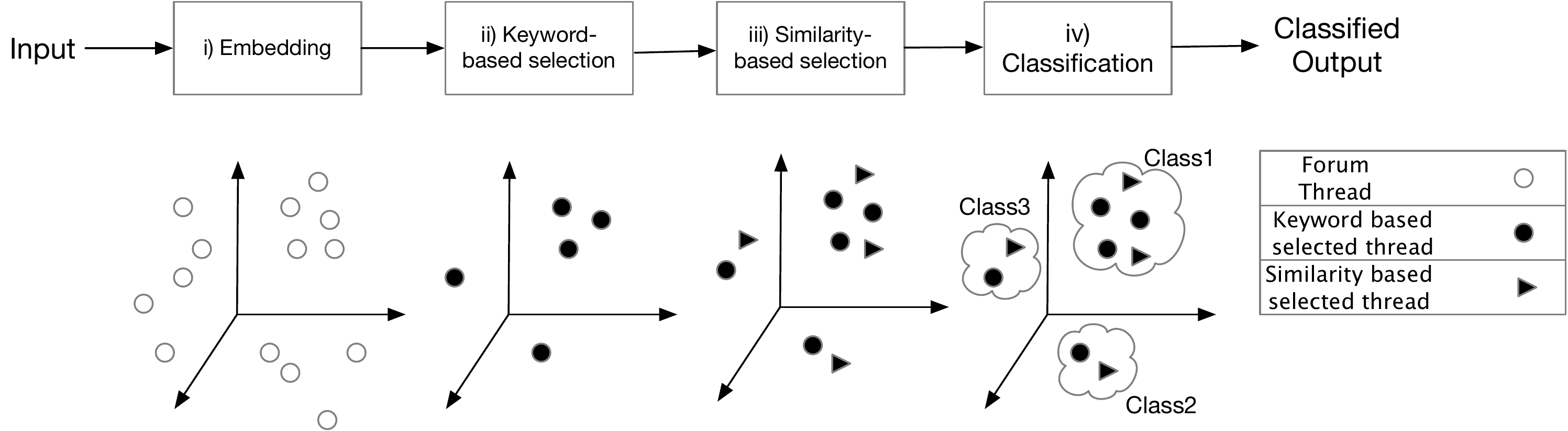}
  \centering
  \caption {An overly-simplified overview of analyzing a forum using the REST approach: i) {\bf project} all threads  to embedding space, ii) {\bf select} relevant threads  using keyword-based selection, iii) {\bf expand} by adding similar threads, iv) {\bf classify} the threads into classes using supervised learning. We illustrate the embedding space as a three dimensional space. }  
  \label{fig:model}
\end{figure*}

Second, we add one more layer of complexity to the problem. To further facilitate the user, we want to group the relevant threads into classes. Again,
 the user defines these classes by providing keywords for each class.
We refer to this step as the {\bf classification} problem.
Note that the user can specify the classes of interest fairly arbitrarily, as long as there is training data
for the supervised-learning classification. 


There is relatively limited work on extracting information from security forums, and even less work on using embedding techniques in analyzing online forum data.   
 We can group prior work in the following categories. First, there is work that analyzes security forums to identify malicious activity~\cite{Portnoff2017,Tavabi2018,Joobin2018}. Moreover, there are some efforts to detect malicious users \cite{Li2014,Marin2018_keyhacker} and emerging threats on forums and other social networks~\cite{Sapienza2017_USC1,Sapienza2018_USC2}. Second, 
 there are several studies on analyzing online forums without a security focus~\cite{Zhang2017,Cong2008SIGIR}.
 Third, there is a large body of work in embedding techniques for: (a) analyzing  text in general~\cite{Mikolov2013,fasttext2017}, and (b) improving text classification~\cite{Bert2018,Wang2018,Tang2015}. 
 Also, note that there exist techniques that can do transfer learning between forums and thus,  eliminate the need to have training data for every forum~\cite{Joobin2018}. 
We discuss related work in more detail in our related work section.




 We propose a systematic approach to identify and classify threads of interest
 based on a multi-step weighted embedding approach. 
 Our approach consists of two parts:
(a) we propose a similarity-based approach with thread embedding to extract  relevant threads reliably, and
b) we propose a  weighted-embedding based classification method to group relevant threads  into user-defined classes.

The key technical foundation of our approach relies on:
(a) building on a word embedding to define thread embedding,
and (b) conducting similarity and classification at the thread embedding level.
Figure~\ref{fig:model} depicts a
 high-level visualization of the key steps of our approach:
  (a) we start with a word embedding space and we define a thread embedding where
 we  project the threads of the forum,
 (b)  we identify relevant threads to the user-provided keywords, 
 (c) we expand this initial set of relevant threads  using thread similarity in the thread embedding,
 (d) we develop a novel  weighted embedding approach to classify threads into the four classes of interest using ensemble learning.
 In particular, we use  similarity between each word in the the forums and representing keywords of each class in order to up-weight the word embedding vectors. Then we use weighted embeddings to train an ensemble classifier using supervised learning.

 
We evaluate the proposed method with three security forums with 163k posts and 21k unique threads. 
The users in these forums seem to have a wide
range of  goals and intentions. 
For the evaluation, we created a labelled dataset of 1350 labeled threads across three forums, which we intend to make available to the research community. 
We provide more information on our datasets in
the next section. 

Our results can be summarized into the following points:


{\bf  a. Providing robustness to initial keyword selection.}
We show that our similarity-based expansion of the user-defined keywords provides significant improvement and stability compared to simple
keyword-based matching.
First, the effect of the initial keyword set is minimized: 
by  going from  240  to  300 keywords,  the keyword-based method identifies 25\% more threads, while the similarity based method increases by only 7\%.
 Second, our approach increases the number of relevant threads 
by 73-309\% depending on the number of keywords.
This suggests that our approach is less sensitive to
omissions of important keywords.

{\bf  b. The relevant threads are 22-25\% of the total threads.} Our approach reduces the amount of threads to 22-25\% of the initial threads. Clearly, these
results will vary depending on the keywords given by the user and the type of the forum. 

{\bf  c. Improved classification accuracy.} Our approach
classifies threads of interest in four different classes with
an accuracy of 63.3-76.9\% and weighted average F1 score 76.8\% and  74.9\% consistently outperforming five other approaches.

{\em Our work in perspective.}
Our work is building block towards a systematic, easy to use,
and effective mining tool for online forums in general.
Although here we focused on security forums, it could easily apply to other forums, and provide the users with
the ability to define topics of interest by providing one or more
set of keywords. We argue that our approach is easy to use
since it is robust and forgiving w.r.t. the initial keyword set.

\section{Definitions and Datasets}
\label{sec:data}

\begin{table}[h]
\centering
\small
 \begin{tabular}{|p{1.1cm}|r|r|r|r} 
 \hline
   & \OffCommShort & \HTS & \Ethical  \\ 
 \hline
  Posts  & 25538 & 84,745 & 54176 \\ 
 \hline
 Users  & 5549 & 5904 & 2970 \\ 
 \hline
 Threads  & 3542 & 8504 & 8745 \\ 
 \hline
\end{tabular}
 \caption {The basic statistics of our forums.}
 \label{tab:forums}
\end{table}

We have collected data from three different forums:
\OffComm, \HTS and \Ethical. 
These forums seem
to bring together a wide range of users:  system administrators,  white-hat hackers,  black-hat hackers, and
users with variable skills, goals and intentions. 
We briefly describe our three forums below.

{\bf a. \OffComm (OC):} This forum seems to be on the fringes of legality. As the name suggests, the forum focuses on ``offensive security", namely, breaking into systems.
Indeed, many posts  provide step by step instructions
on how to compromise systems, and  advertise hacking tools  and services.

{\bf b. \HTS (HT):} As the name suggests, this forum  has also an attacking orientation. There are threads
that describe how to break into websites and systems,  but there are also more general discussions about the users' experiences in cyber-security. 

{\bf \Ethical (EH):} This forum  seems to consist mostly of  ``white hat" hackers, as its name suggests. Many threads are about making systems more secure. However,  there are many discussions with malicious intents are going on in this forum. Moreover, there are some notification discussions to alert about emerging threats.

\begin{figure}[!htb]
\begin{minipage}{0.32\linewidth}
  \includegraphics[width=\linewidth]{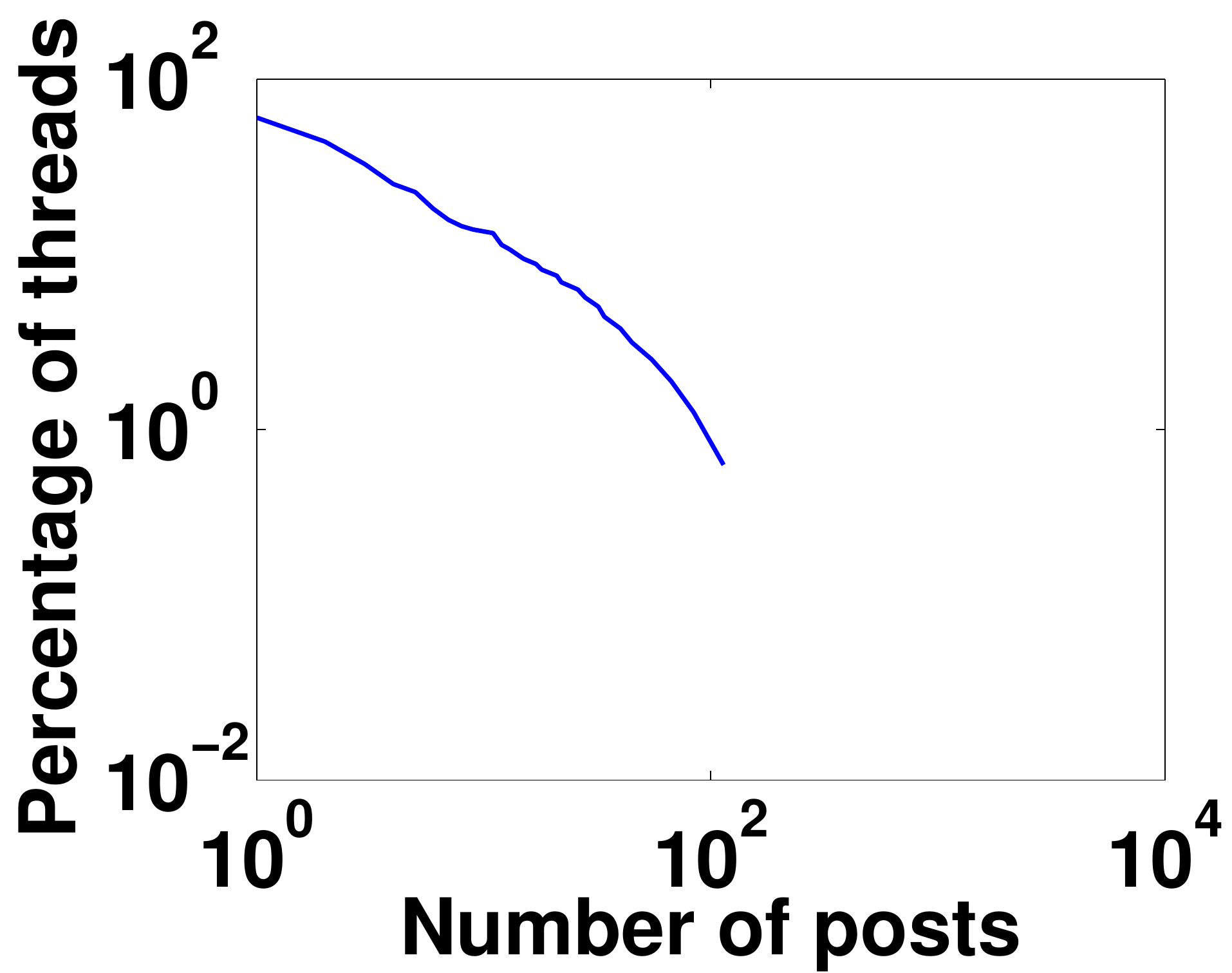}
  \centering
  {(a)\OffCommShort}
\end{minipage}
\begin{minipage}{0.32\linewidth}
  \includegraphics[width=\linewidth]{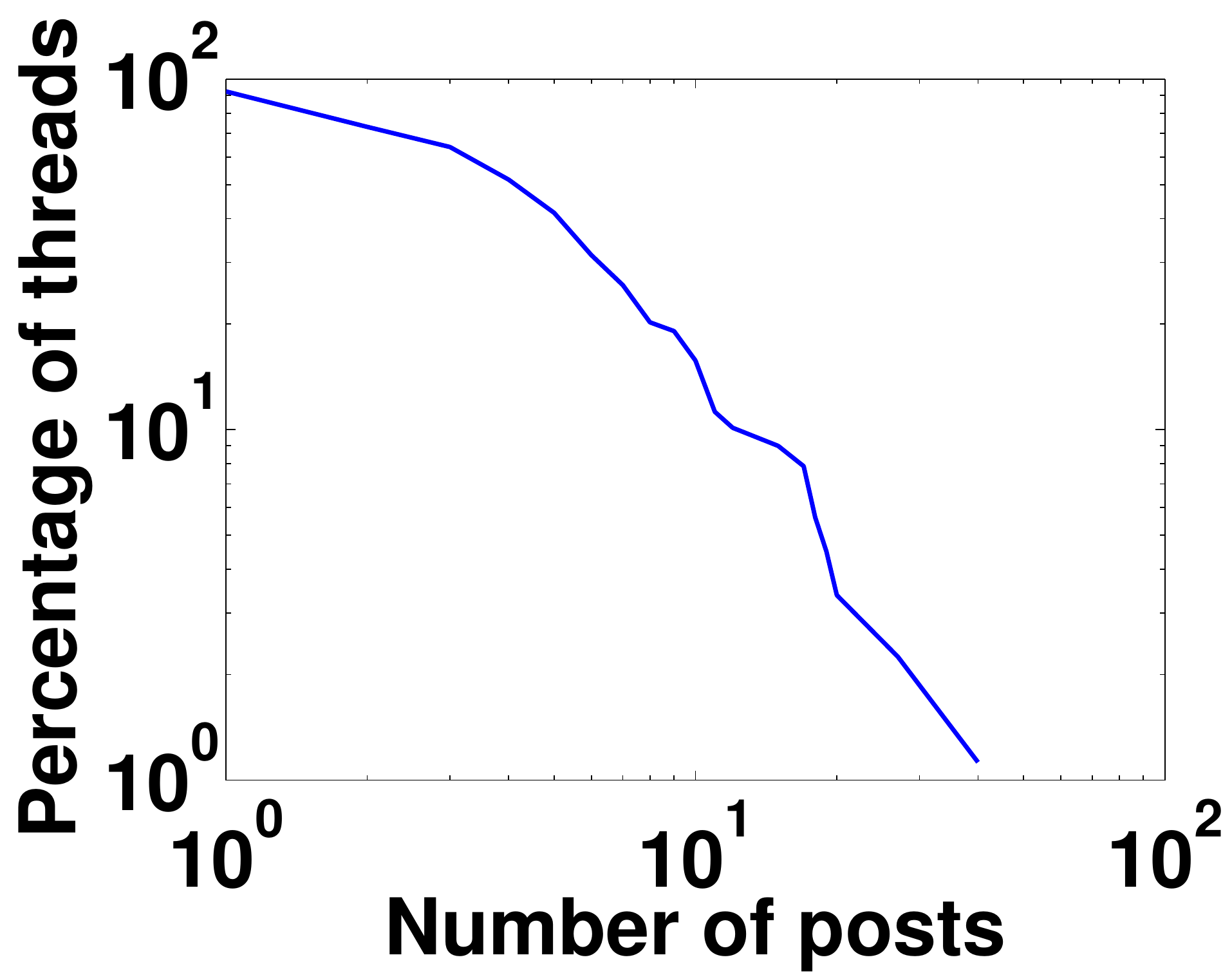}
  \centering
  {(b)\HTS}
\end{minipage}
\begin{minipage}{0.32\linewidth}%
  \includegraphics[width=\linewidth]{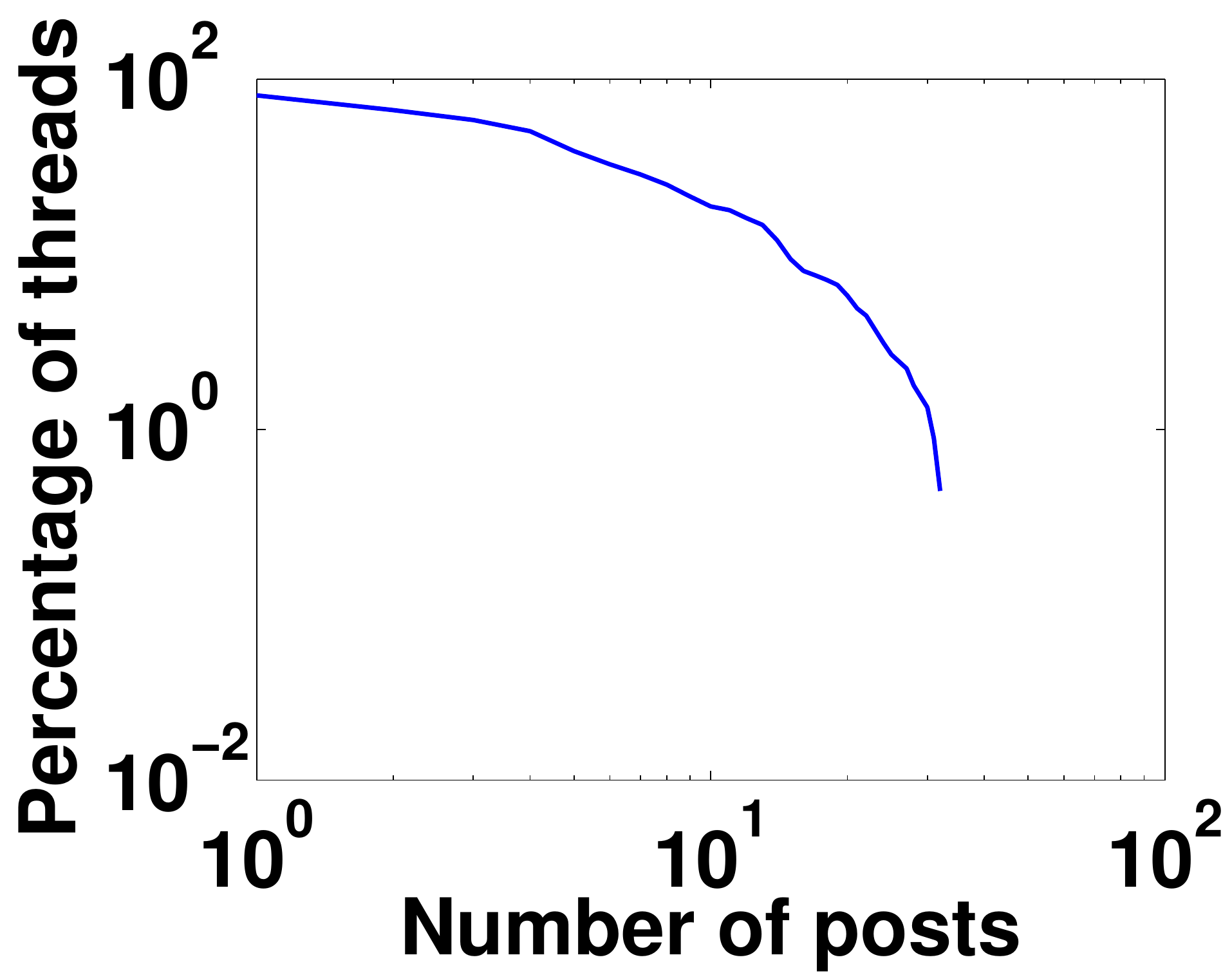}
  \centering
  {(c)\Ethical}
\end{minipage}
\caption{CCDF of the number of Post  per thread ($\log$-$\log$ scale).} 
\label{fig:ccdf}
\end{figure}

{\bf Basic forum statistics.}
We present basic statistics of our forums in Table \ref{tab:forums}. We also study some of their properties
and make the following two observations.

{\bf Observation 1: More than half of the threads have one post!} In Figure \ref{fig:ccdf}, we plot the complementary cumulative distribution function of the number of post per thread for our forums. 
We observe the skewed distribution that describes the behavior
of large systems. In addition, the distribution shows that more than half of thread has one single post in the threads and 73\% of the threads has one or two posts in threads.

{\bf Observation 2: The first post defines the thread.}
Prior research~\cite{Zhang2017} seems to confirm something that we intuitively expect: 
the first post of the thread pretty much defines the thread.
 Intrigued, we sampled and  manually verified that this seems to be the case. Specifically, we inspected a random sample of 10\% of the relevant threads (found by our approach),
 and we found that more than 97\% of the follow up posts fall
 in line with the topic of the thread: while a majority of them,  express appreciation,  agreement etc. 
   For example, the follow up posts to a malicious tutorial in \OffComm were: ``Great Tut", ``Thank you for sharing", ``Nice post", ``Work[s] great for me!"
 
 

{\bf Defining the classes of interest.}
As we explained in the introduction, we want to further help a security analyst by giving them the ability to define classes of interest among the threads of interest. These are user-defined classes. To ground our study, we focus on the following classes,
which we argue could be of interest to a security analyst.

{\bf a. \AN :} These are  threads where users are reporting about being attacked by a hackers or notifying about exploits and vulnerabilities. An example from \Ethical is a thread with the title ``Worm Hits Unsecured Space Station Laptops" and the first line of the first post is ``NASA spokesman Kelly Humphries said in a statement that this was not the first time that the ISS had been affected by malware, merely calling it a “nuisance.”"

{\bf b. \PS:} These are  threads where users are offering or requesting malicious hacking services or products.  
An example from \OffComm is a thread with the title ``Need hacking services'' and this first line ``Im new to this website. Im not a hacker. Would like to hire hacking services to hack email account, Facebook account and if possible iPhone.''

{\bf c. \TT:} These are threads where  users post detailed instructions for performing malicious activities. The difference with the above category is that the information is offered for free here. An example from \OffComm is a  thread titled 
``Hack admin account in XP, Vista, Windows 7 and Mac - Complete beginners guide!!" with a first line: ``Hack administrator account in XP OS – Just by using command prompt is one of the easiest ways (without installation of any programs).....". 
As expected, these posts are often lengthy as they convey detailed information. 

{\bf d. \AG:} These are threads  where users share their experience related to  general security topics. Often users provide a personal story, a review or an article on a cyber-security concept or event. For example, in \HTS  a thread titled  ``Stupid people stories", the author explains
 cyber-security mistakes that he made.
 
 The sets
 of  keywords which ``define" each class are shown in Table \ref{tab:keywords}. Clearly, these sets will be provided by the user depending on classes of interest. Note that these keywords are also provided to our annotators as hints for labeling process. 
 

\subsection{Establishing the Groundtruth}

\begin{table}[t]
\centering
\small
 \begin{tabular}{|l|r|r|r|r|r|r|} 
 \hline
   & \multicolumn{2}{|c|}{\OffCommShort} & \multicolumn{2}{|c|}{\HTS} & \multicolumn{2}{|c|}{\Ethical}  \\ 
 \hline
 Labeled  & \multicolumn{2}{|c|}{450} & \multicolumn{2}{|c|}{450} & \multicolumn{2}{|c|}{450}\\ 
 \hline \hline
    & \# & \%  & \# & \% & \# & \% \\
 \hline
 \TT   & 202 &45\%  & 31 &7\% & 42 &9\% \\
 \hline
 \PS  & 204 &45\% & 286 &64\% & 166 &37\%\\ 
 \hline 
 \AN  & 27 &6\% & 20 &4\% & 78 &18\%\\
 \hline
 \AG & 17 &4\% & 113 &25\% & 164 &36\% \\ 
 \hline
\end{tabular}
 \caption { Our groundtruth data for each forum and the breakdown per class.    \label{tab:labels} } 
\end{table}

For validating our classification method, we need groundtruth to do both the training and the validation. We randomly selected 450 threads among the relevant threads from each forum as selected by the identification part.
The labelling involves  five annotators that manually label each thread to a category based on the definitions and examples of the four classes which we listed above. The annotators were selected from a educated and technically savvy group of individuals to improve the quality of the labels.
We then combine the ``votes", and assign the class selected by the majority.


We assess the annotators' agreement based on the Fleiss-Kappa coefficient and we show the results in Table~\ref{tab:kohenCoeff}. We see that there is a high annotator agreement across all forums as the Fleiss-Kappa coefficient is 78.6. 92.6, 70.3 for \OffComm, \HTS and \Ethical respectively. 


With this process, we labelled 1350 posts in three forums
and we present our labeled data in Table~\ref{tab:labels}.
We make our groundtruth available to the researchers in the community in order to foster follow up research.
\footnote{Data is provided  at the following link: https://github.com/icwsmREST2019/RESTDATA.}
\begin{table}[t]
\centering
\small
 \begin{tabular}{|l|r|r|r|r|} 
 \hline
 Label  & \TT & \PS & \AN & \AG  \\
 \hline
 \OffCommShort & 0.778 &	0.702 &	0.816 & 0.732\\ 
 \hline 
 \HTS & 0.953 &	0.966 &	0.793 & 0.875\\
 \hline
 \Ethical  & 0.682 &	0.733 &	0.766 &	0.620  \\ 
 \hline
\end{tabular}
 \caption { Assessing the annotator agreement using the Fleiss-Kappa coefficient for each class for our three datasets.    \label{tab:kohenCoeff} } 
\end{table}


\subsection{Challenges of simple keyword-based filtering}

Given a set of keywords,
the most straightforward approach in identifying relevant documents (or threads here) is to count the combined frequency with which these keywords appear in the document.
A user needs to identify the keywords that best describe the topics and concepts of interest, which can be
challenging for non-trivial scenarios~\cite{Wang2016}.
We outline some of the challenges below.

\begin{itemize}
    \item The user may not be able to provide all keywords of interest.
    In some cases,
    the user is not aware of a term, and in some cases, this not even possible: consider the case where we want to find the name of a new malware that has not yet emerged. 
    
    \item Stemming, variations and compound words is a concern. The root of a word can appear in many different versions: e.g. hackers, hacking, hacked hackable, etc.  
    There exist partial solutions for stem but
    challenges still remain~\cite{stemming2011}.
    
    \item Spelling errors and modifications and linguistic variations. Especially for an international forum, different languages and
    backgrounds can add  noise.
    
    
\end{itemize}

The above challenges motivated us to consider a new approach that uses a small number of indicative keywords to create a seed set of threads, and then use similarity in the embedding space to find more similar threads, as we describe in the next section.
\section{Identifying threads of interest}
\label{sec:identify}

\begin{table}[t]
\centering
\small
 \begin{tabular}{c|c} 
 \hline
  Symbol & Description \\
 \hline
 \hline 
 $\word_i$ & Word $i$ in a forum\\
 $\vec{v_i}$   & Embedded vector for word $i$   \\
 $\word_{i,k}$ & Value of dimension $k$ in embedded vector for word $i$ \\
 $t_\tidx$ & Thread $\tidx$ \\
 $\mdim$  & The dimensions of the word embedding space \\
 $n$ & Number of words in a thread\\
 $d$ & Number of words in a forum \\
 $\Words(t_\tidx)$ & Set of words in thread $\tidx$ \\
 $D$ & Set of words in a forum \\
 $\vec{\pro}(e)$ & Embedding projection of entity $e$ (word, thread etc) \\
 $Sim(w,c)$ & Similarity of vectors $w$ and $c$\\
  $\classw_\classi$ & The ``center of gravity" word for class $\classi$ \\
 $\vec{\beta}_\classi $ & Affinity vector of class $\classi$ \\
 $\vec{\beta}_\classi[i] $ & Value of Affinity vector of class $\classi$ at index $i$ \\
  $WS_l$ & Keyword set l for identifying relevant threads\\
 $\ThWord$ & Keyword threshold in identifying relevant threads\\
 $\ThSim$ & Similarity threshold in identifying relevant threads\\
   \hline
 \hline
\end{tabular}
 \caption { The symbol table with the key notations.}
 \label{tab:def}
\end{table}

We present our approach for selecting relevant threads starting from sets of keywords provided by the user.
Our approach consists of the following phases: 
(a) a  keyword matching step, where we use the user-defined keywords to identify relevant threads that contain these keywords,
and (b) a similarity-based phase, where we identify threads that are ``similar" to the ones identified above.
The similarity is established at the word embedding space as we describe later.

\subsection{Phase 1: Keyword-based selection}

Given a set or sets of keywords, we identify the threads 
where these keywords appear.  A simple text matching approach can distinguish all occurrence of such keywords in the forum threads. In more detail, we follow the steps below:

 {\bf Step 1:} The user provide a set or sets of keywords $WS_l$, which  capture the user's topics of interest. Having sets of keywords enables the user to specify combinations of concepts. For example, in our case we use,
 the following sets: (a) hacking related, (b) exhibiting concern and agitation, and (c) searching and questioning.

{\bf Step 2:} We  count the frequency  of each keyword in all the threads. This can be done easily with elastic search or any other straightforward implementation.

{\bf Step 3:} We identify the relevant threads, as the threads that contain a sufficient number of keywords from each set of keywords $\WS_l$.
This can be defined by a threshold, $T_{key_l}$, for each set of keywords.

 Going beyond simple thresholds in this space, we envision a flexible system,
 where the user can specify complex queries that involve combinations of several different keyword sets $\WS_l$. For example, the user may want to find threads with: 
 (a) at least 5 keywords from set $\WS_1$ and 3 keywords from $\WS_2$,
 or (b) at least 10 keywords from $\WS_3$.



\subsection{Phase 2: Similarity-based selection}

 We propose an approach to extract additional relevant threads based on their similarity to existing relevant threads. Our approach is inspired by and combines elements from earlier 
 approachs~\cite{Mikolov2013,Shen2018},
 which we discuss and contrast with our work in the related work section.

  {\bf Overview of our approach.} Our approach follows the  steps below,
 which are also depicted visually in Figure~\ref{fig:model}.
 The input is a forum, a set of keywords, and set of relevant threads, as identified by the keyword-based phase above.
 
 {\bf Step 1. Determining the embedding space.} We project every word as a point in a $\mdim$-dimensional space using a word embedding approach. Therefore, every word is represented by a vector of $\mdim$ dimensions.
 
{\bf Step 2. Projecting threads.} We project all the threads in  an appropriately constructed multi-dimensional space: both the relevant threads selected from the keyword-based selection and the non-selected ones. The thread projection is derived from the projections of its words, as we describe below. 
   
{\bf Step 3. Identifying relevant threads.} We  identify more relevant threads among the non-selected threads that are ``sufficiently-close" to the relevant threads in the thread embedding space. 

The advantage of using similarity at the level of threads is that
thread similarity can detect high-order levels of similarity, 
beyond keyword-matching.
Thus, we can identify threads that do not necessary exhibit the keywords,
but use other words for the same ``concept".
We show examples of that in Tables~\ref{tab:SimilSample} and~ \ref{tab:SimilSample2}.

    

{\bf Our similarity-base selection in depth.}
We provide some details in several aspects of our approach.

{\bf Step 1: in depth.} We train a
 skip-gram word embedding model to project every word as a vector in a multi-dimensional space~\cite{Mikolov2013}. Note
 that we could not use pre-trained embedding models,
 since there are many words in our corpus that do not exist in the dictionary of previous models.

The number of dimensions of the word embedding can be specified  by the user:  NLP studies usually opt for a few hundred dimensions. 
We discuss how we selected our dimensions in our experiments section. 

At the end of this step, every word $\word_i$ is projected to  $\vec{\pro}(\word_i)$ or $\vec{\word}_i$, 
a real-value $\mdim$-dimensional vector,
$(\word_i[1],\word_i[2], ...,\word_i[\mdim])$.
A good embedding ensures that two words are similar,  if they are close in the embedding space. 
 
{\bf Step 2: in depth.}
 We project the threads in an $2\mdim$-space, by ``doubling" the $\mdim$-dimensional space that we used for words as we will show below.
 The thread projection is a function of the vectors of its words
 and captures both the average 
 and the maximum values of the vectors of its words. 
   
   \indent {\bf a. Capturing the average: $\pro_{avg}(t_\tidx)$.} Here, we want to capture the average ``values" of the vectors of the words in the thread. 
   For thread $t_\tidx$, the average projection, $\pro_{avg}(t_\tidx)$  is calculated as follows
   for each dimension $l$ in the $\mdim$-dimensional word space:
   \begin{equation} \label{eq:avg}
     \vec{\pro}_{avg}(t_\tidx)[l] = 
         \frac{1}{|\Words(t_\tidx)|} \cdot 
         \sum_{\word_i \in  \Words(t_\tidx)}^{} \vec{v}_i[l] , 
\end{equation}

Recall that $\Words(t_\tidx)$ is the set of words of the thread.
For simplicity, we refer to projection of word $\word_i$ as $\vec{\word}_i$ instead of the more complex $\vec{\pro}(\word_i)$.
    
    \indent {\bf b. Capturing the high values: $\pro_{max}(t_\tidx)$.}
    Averaging can fail to represent adequately
    the ``outlier" values, and to overcome 
    this, we calculate a vector of maximum values, $\vec{\pro}_{max}(t_\tidx)$, for each thread.
    For each dimension $l$ in the  word embedding,
     $\pro_{max}[l]$ is the maximum value of that dimension over all existing $l$-dimension values among all the words in the thread, which we can state more formally below:
    
    \begin{equation} \label{eq:max}
        \vec{\pro}_{max}(t_\tidx)[l] = \max_{\word_i \in  \Words(t_\tidx)} \vec{v_{i}}[l]
    \end{equation}

Finally, we create the projection of thread $t_\tidx$ by using both these vectors,  $\vec{\pro}_{avg}(t_\tidx)$ and $\vec{\pro}_{max}(t_\tidx)$,
    as this combination has been shown to provide good results~\cite{Shen2018}. Specifically, we concatenate the vectors
    and we create the thread representations in an $2\mdim$-dimensional space.
    
    \begin{equation} \label{eq:concat}
        \vec{\pro}(t_\tidx) = (\vec{\pro}_{avg}(t_\tidx) , \vec{\pro}_{max}(t_\tidx))
    \end{equation}

{\bf Step 3: in depth.}
    We  identify similar threads at the   $2\mdim$-space-dimensional space of thread embedding from step 2. 
    We propose to use the cosine-similarity determine the similarity among threads, which seems to give good results in practice.
    Most importantly, we can control what constitutes a 
    {\em sufficiently-similar} thread using a threshold $\ThSim$.
    The threshold needs to strike a balance between being too selective and too loose in its definition of similarity. Furthermore, note the 
    right threshold value also depends on the nature of the problem and the user preferences. For example, a user may want to be very selective, if the resources for further analyzing relevant threads is limited or if
    the starting number of threads is large.

\section{Classifying threads of interest}
\label{sec:classification}

\begin{figure}[htb]
  \includegraphics[scale=0.4]{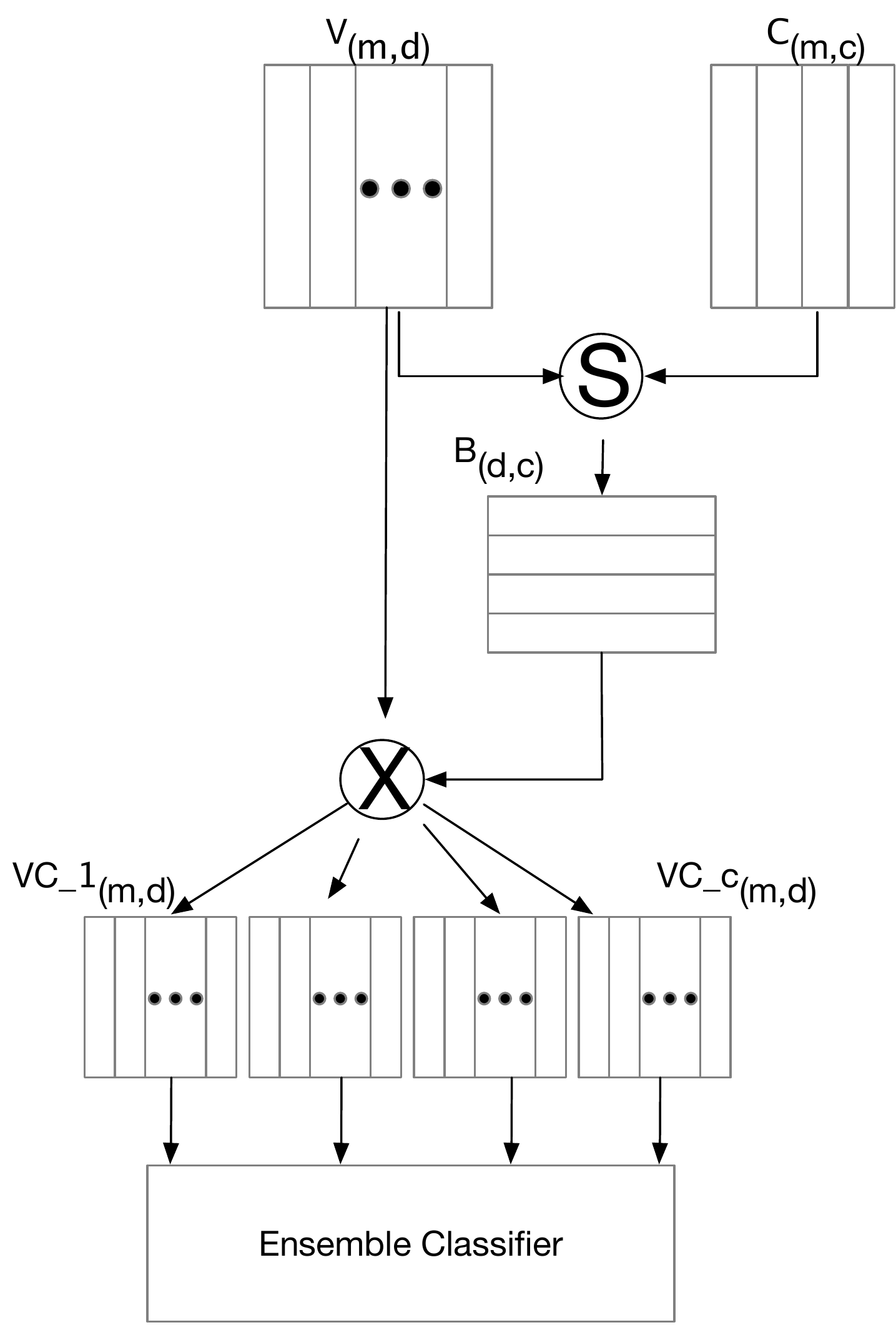}
  \centering
  \caption {A visual overview of our classifier }
  \label{fig:classifier}
\end{figure}

We present our approach for classifying relevant threads into user defined classes. To ground the discussion, we presented four classes on which
we focus here, but our approach can be applied for any number and type of classes as long as there is training data for the supervised learning.

{\bf Defining Affinity.}
We use the term {\bf affinity}, $\vec{\beta}_\classi[i]$ , to refer to the 
``contribution" of word $\word_i$ in a thread towards our belief that the thread belongs to class $\classi$.

Recall also that each class $\classi$ is characterized by a group of words that we denote as $\WordClass_\classi$.
These sets of words are an input to our algorithm and in practice they will be provided by the user.

{\bf High-level overview creating our classifier.} 
Our approach consists of the following steps, which are visually represented in Figure~\ref{fig:classifier}.

{\bf Step 1.} We create a representation of every class $\classi$ into the word embedding space by using the words that define the class, $\WordClass_\classi$. 



{\bf Step 2.} 
For all the words in the forum,
we calculate the affinity of the word $\word_i$ for each class $\classi$, $\vec{\beta}_\classi[i]$.  

{\bf Step 3.}
For each class, we create a weighted embedding by using 
the affinity to adjust the embedding projection of each word 
for each class.

{\bf Step 4.} We use weighted embedding to train an ensemble classifier using supervised learning. 

{\bf Using the classifier.} Given a thread, we calculate its projection in the embedding space, and then we pass it to the classifier to determine its class.




{\bf Our algorithm in more detail.}
In the remainder of this section, we provide a more in depth description of  the algorithm.

{\bf Step 1: in depth.} For each class $\classi$, 
we use the set of words $\WordClass(\classi)$, and to define a  representation, $\vec{\classw}_\classi$, for that class in the word embedding space. 
We project each word in $\WordClass(\classi)$ to the embedding space by using the same word embedding model, which we trained in the previous section. 
Then, we define the class vector $\classw_\classi$ to be the average of the  word embeddings  of the words in $\WordClass(\classi)$  similarly to  equation~\ref{eq:avg}. Note that these 
class embedding vectors correspond to each column of the matrix $C_{(\mdim, c)}$ in Figure \ref{fig:classifier}, where $\mdim$ in the dimension of the embedding and $c$ is the number of classes.

{\bf Step 2: in depth.}
The affinity of each word  $\word_i$ in the forum for each class is calculated 
by the similarity of the word $\word_i$ to  $\vec{\classw}_\classi$, which represents the class in this space. 
We calculate the proximity using the cosine similarity, as follows: 

\begin{equation} \label{eq:sim}
   Sim(\vec{\word_i},\vec{\classw}_\classi) = \frac{\vec{\word}_i \cdot \vec{\classw}_\classi}
                     {||\vec{\word_i}|| \cdot || \vec{\classw}_\classi||} 
\end{equation}

Then, for each class $\classi$, we  create vector $\vec{\beta_\classi}$ whose element $[i]$ corresponds to the affinity of word $\word_i$ of the forum $D$.
Specifically, 
we normalize the values by using \textit{Softmax} of the similarity vector  $Sim(\word_i,w_\classi)$  as follows:

\begin{equation} \label{eq:softmax}
   \vec{\beta_\classi}[i] = \frac{exp(Sim(\vec{\word}_i, \vec{\classw}_\classi))}
                   {\sum_{y_j \in \Dict}^{}exp(Sim(\vec{y}_j, \vec{\classw}_\classi))}  \space ,
\end{equation}


where $y_j \in \Dict$ iterates through all the words in the forum.
Note that $\vec{\beta_\classi}$ corresponds to a row $\classi$ in matrix $B_{d,c}$ in figure \ref{fig:classifier}, where $c$ is the number of classes and $d$ is the total number of words in the forum.

{\bf Step 3: in depth.} For each class $\classi$, we create
a ``custom" word embedding, $VC_\classi(m,d)$ in Figure \ref{fig:classifier}. Each such matrix
that is focused on detecting threads of $\classi$
and it will be used in our ensemble classification.
  
  For each class, we create, $VC_\classi(m,d)$, a class-specific word embedding by modifying  the word projections, $\vec{\word}_i$ using the affinity of the word $\vec{\beta}_\classi[i]$ for class. Formally, we calculate $VC_\classi$ by calculating  column $VC_\classi[ *, i]$ as  follows: 
 
 \begin{equation} \label{eq:VC}
    VC_\classi[ *, i] = \vec{\beta_\classi}[i] \cdot \vec{\word}_i  
 \end{equation}
 
 where $\vec{\beta}_\classi[i]$ is  the affinity value of word $\word_i$ for class \classi.
  
  For each thread $t_\tidx$, we calculate the projection of the thread by calculating $\vec{\pro}_{avg}(t_\tidx)$ and  $\vec{\pro}_{max}(t_\tidx)$ 
  using the modified word projections, 
  $\vec{\beta_\classi}[i] \cdot \vec{\word}_i$, captured in the $VC_\classi(m,d)$ matrix and using
   equations \ref{eq:avg} and \ref{eq:max}.
 Finally, we create the projection of each thread in the $2\mdim$-space, using equation~\ref{eq:concat}.


{\bf Step 4: in depth:} We use weighted embeddings of threads to train an ensemble classifier using supervised learning. 

For each class $\classi$, we train the classifier by using the weighted representation vector in a supervised learning. 
Each $VC_\classi$ in Figure \ref{fig:classifier}
becomes the basis for a classifier  with weighted penalty in favor of class $\classi$.
The ensemble classifier combines the classification results from 
each $VC_\classi$ classifier using the max-voting approach~\cite{maxvoting1998}.

{\bf Using contextual features.}
Apart from the words in the forum, we can also consider other types of features, which we refer to as contextual features of the threads. One could think of various such features, but here we list the features that
we use in our evaluation: (1) number of newlines, (2) length of the text, (3) number of replies in the thread (following posts after the first post), (4) average number of newlines in replies, (5) average length of replies, and (6) the aggregated frequency of the words of each bag-of-words  set provided by the user. 

These features capture contextual properties of the posts in the threads,
and provide additional information not necessarily captured by the words
in the thread. 
Empirically, we find that these features improve the classification accuracy significantly. The inspiration to introduce such features came from  manually inspection of posts and threads. For example, we observed that \TT and \AG usually have longer posts than other. Moreover, \TT threads contain a larger number of newline characters. 
An interesting question is to assess the value of such metrics when
used in conjuction with word-based features.




\section{Experimental Results}

\begin{figure}[t]
  \includegraphics[width=0.7\linewidth,scale=0.25]{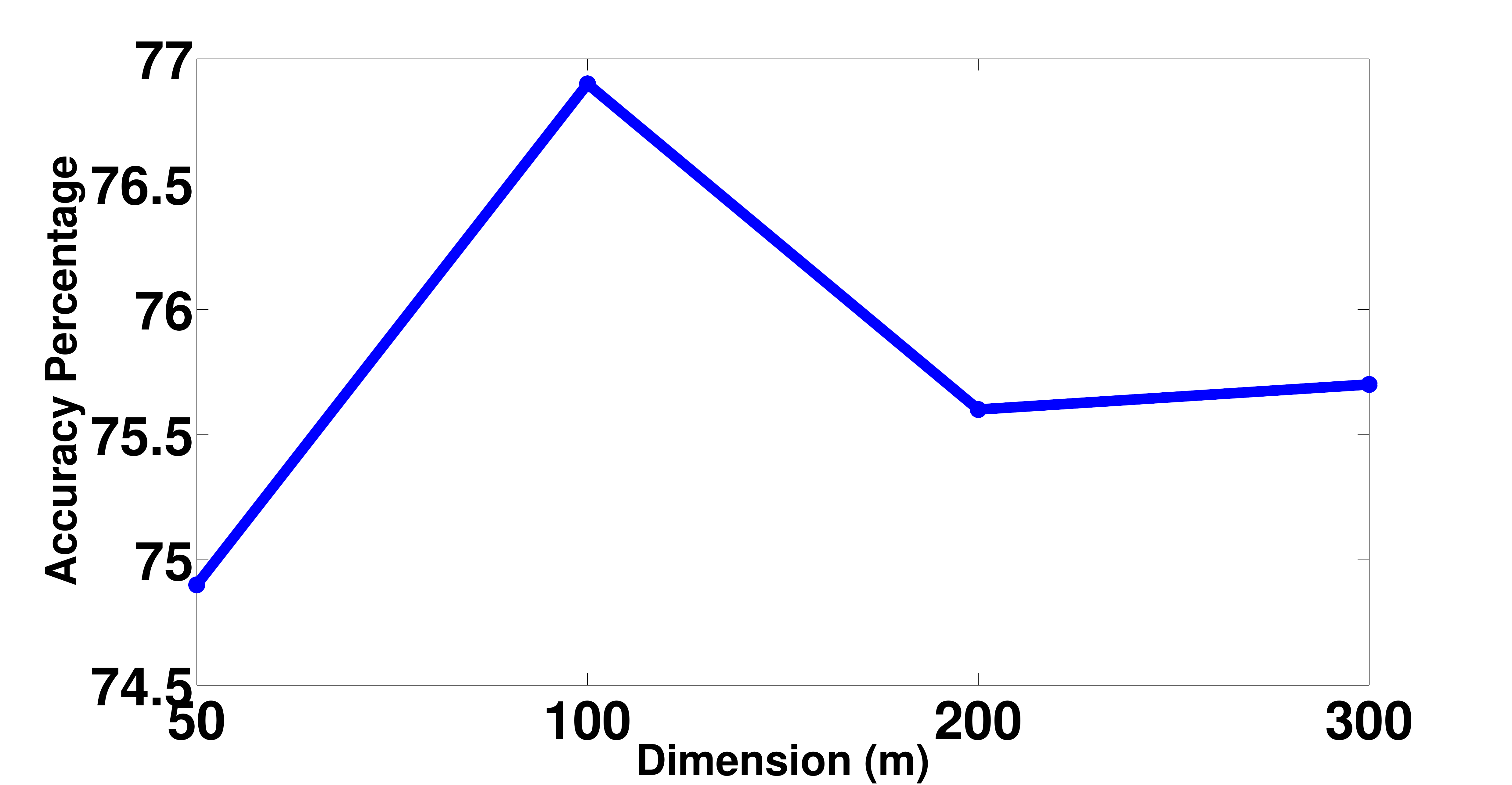}
  \centering
  \caption { Selecting the number of dimensions of word embedding: 
  the accuracy of \myalg for different dimensions in \OffComm.}
  \label{fig:bestM}
\end{figure}


\begin{table}[t]
\centering
\small
 \begin{tabular}{|c|c|c|c|} 
 \hline
  \TT & \PS & \AN & \AG  \\ 
 \hline \hline
  Tutorial & tool & announced & article \\
  guide & price & reported & story  \\
  steps & pay & hacked & challenge \\
 \hline
\end{tabular}
\caption {$\WordClass$, the set of words which "define" each class.} 
\label{tab:keywords}
\end{table}

We present our experimental results and evaluation of our approach.

\subsection{Conducting our study}

We use the three forums that presented in Table~\ref{tab:forums} and the groundtruth, which we created as we explained in section definitions.

{\bf Keywords sets}: We considered three keyword sets to capture relevant threads. These keywords set are: (a) hacking related, (b) exhibiting concern and agitation, and (c) searching and questioning. We collected a set of more than 290 keywords in three sets. 
We started with a small core group of keywords,
which we expanded by adding their synonyms using thesaurus.com and Google's dictionary. 
We ended up with 68, 207 and 17 keywords for the three groups respectively.

These keyword sets are used in extracting relevant threads with the keyword-based selection. We select a thread, if it contains at least one word from each keyword set: $T_{key_1},T_{key_2},T_{key_3} >=1 $. 
As we discussed earlier, there are many different ways to perform
this selection in the presence of multiple groups of words and depending
on the needs of the problem.

{\bf Pre-processing text:} As with any NLP method, we do several pre-processing steps  in order to extract an appropriate set of words from the target document.  First we tokenize the documents in to bigrams, then we remove the stopwords, numbers and IP addresses based on a recent work~\cite{Joobin2018}.
In addition,
here we opt to focus on the title and the first post of a threads instead of using all the posts. Our rationale is based on the two observations regarding the nature of the threads: (a) most of them have one post anyway,
and (b) the title and the post typically define their essence.
In the future, we will examine the effect of using the text from more posts from each thread.

{\bf Identification: Implementation choices.} 
The identification algorithm requires some implementation choices, which we describe below.

{\bf Embedding parameters:} We set the window size to 10 and we tried several different values as the dimension of the embedding between 50-300, and we found that $\mdim = 100$ with the highest accuracy as depicted in Figure \ref{fig:bestM} and
\mdim is in the range of choice of other studies in this space.  

{\bf Similarity threshold: $\ThSim = 0.96$.} The similarity threshold $\ThSim$ determines the ``selectiveness" in identifying similar threads, as we described in a previous section. We find that a value of 0.96 
 worked best among all the different values we tried.
 It  strikes  the  balance between being: sufficiently selective  to filter out non-relevant threads, but sufficiently flexible to identify similar threads.  

 

{\bf Classification: Implementation choices.} 
We present the implementation choices for our classification study.

{\bf Evaluation Metrics:} We used the accuracy of the classification  along with the average weighted  F1 score, which is designed to
take into consideration the size of the different classes in the data.

{\bf Our classifier.} We use random forest as our classification engine, which performed better than several others that we examined, including SVM, Neural Networks, and K-nearest-neighbors. Results are not shown due to space limitations.

{\bf Class defining words:} The set of keywords we have used for each class are as shown in Table \ref{tab:keywords}.

{\bf Baseline methods.}
We evaluate our approach against five other state of the arts methods, which we briefly describe below. 

\begin{itemize}
    \item {\bf Bag of Words (BOW)}: This methods  uses the word frequency (more accurately the TFIDF value)  as its main feature \cite{mccallum1998naive,Joobin2017,Jin2016}.
    
    \item {\bf Non-negative Matrix Factorization (NMF)}:
     This method uses linear-algebra to represent high-dimensional data into low-dimensional space, in an effort to capture latent features of the data~\cite{Lee1999}.
    
    \item {\bf Simple Word Embedding Method (SWEM)}: There is a family of methods that use the word2vec as their basis, and  use a recently proposed method \cite{Shen2018}.
    
    \item {\bf FastText (FT)}:
    Similar to NMF and SWEM, FastText
    represents words and text in a low dimensional space~\cite{fasttext2017}. 
    
    \item {\bf Label Embedding Attentive Model (LEAM)}: This is the most recent approach \cite{Wang2018}  claims to outperform other state of art methods including PTE~\cite{Tang2015}. We used their provided linear implementation of their attentive model. 
    
    \item 
     {\bf Bidirectional Encoder Representations from Transformers (BERT)} : This is a new pre-trained Deep Bidirectional Transformer for Language Understanding introduced by Google~\cite{Bert2018}. BERT  provides contextual representation for text, which can be used for a wide range of NLP tasks.
     As we discuss later, BERT did not provide good results initially, and we created a tuned version 
     to make the comparison more meaningful.
\end{itemize}

\subsection{ Results 1: Thread Identification}

\begin{figure}[t]
  \includegraphics[width=1\linewidth]{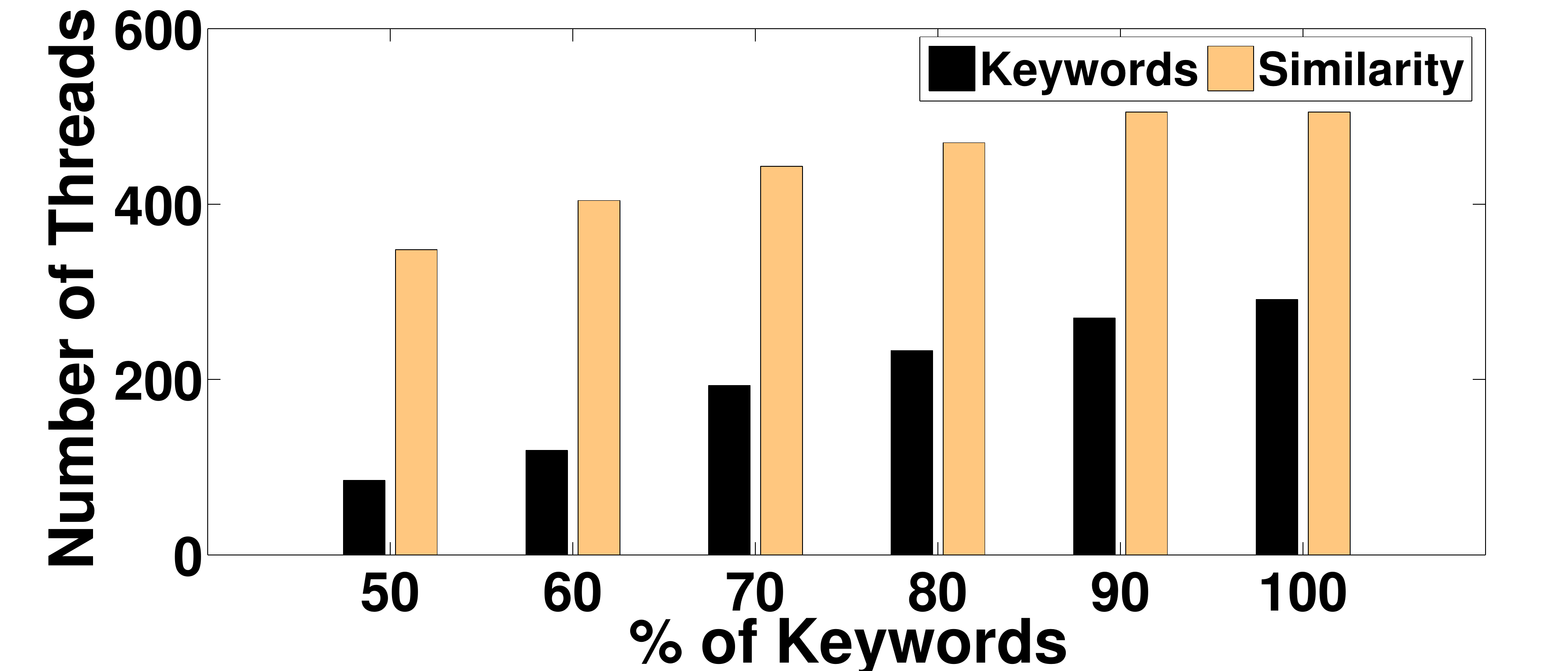}
  \centering
  \caption {The robustness of the similarity approach to the initial keywords: number of relevant threads as a function of the number of keywords for \OffComm.}
  \label{fig:sensitivity}
\end{figure}

\begin{table}[t]
\centering
\small
 \begin{tabular}{|p{1.5cm}|r|r|r|r} 
 \hline
  Relev. Threads & \OffCommShort & \HTS & \EthicalShort  \\ 
 \hline \hline
 Keyword  & 291 & 840 & 893 \\ 
 \hline
 Similarity   & 505  & 1121  & 1360  \\
 \hline
  Total    & 796  & 1961  & 2753  \\
  \hline
  Total(\%)    & 22\%  & 23\%  & 25\%  \\
 \hline
\end{tabular}
\caption {The  relevant threads and their identification method: keywords and similarity. The total percentage refers to the selected threads over all the threads in the forum.} 
\label{tab:relevant}
\end{table}
We present the results from the identification part of our approach.

{\bf  Our similarity-based method is robust to the number of initial keywords.}
We want to evaluate the impact of the number of keywords to the similarity based method. In Figure \ref{fig:sensitivity}, we  show the robustness of
each identification methods to the initial set of keywords for \OffComm . By adding 60 keywords, from 240 to 300, the keyword-based method identifies 25\% more threads, while the similarity based method has only 7\% increment. Similarly, doubling the initial size of the keywords results in 242\% increase for the keyword-based method but only 45\% in the similarity-based method.

We argue that our approach is more robust to the initial number of keywords. First, with less number of keywords, we retrieve more threads. 
Second, an increase in the number of keywords has less relative increase in the number of threads. This is an initial indication that our approach can achieve good results, even with a small initial set of keywords.



{\bf Evaluation of our approach: High precision and reasonable recall.}
We show that
our  approach is effective in identifying  relevant threads. 
Evaluating precision and recall would have been easy if all the threads in a forum were labelled.
Instead, we use an indirect method to gauge recall and precision as we describe below.

{\bf Indirect estimation of recall.}
We consider as ``groundtruth"  the  relevant threads that we find with set of keywords in keyword-based selection method and report how many of those threads that our method finds with only 50\% of the keywords in similarity-based selection. The experiment is  shown in Figure \ref{fig:sensitivity}. We use only 50\% of the keywords to extract the relevant threads with the similarity selection approach, and then compare it with the relevant threads identified with larger set of keywords [60-100]\%.
We show in Table \ref{tab:relevantRecall} that  with 50\% of the keywords we can identify more than 60-70\% of the relevant threads, which we identify if we have more keywords available.

\begin{table}[t]
\centering
\small
 \begin{tabular}{|p{1.6cm}|r|r|r|r|r||r|} 
 \hline
  keywords \% & 60 & 70 & 80 & 90 & 100 & Avg.  \\ 
 \hline 
 \OffCommShort  & 78.2 & 76.9 & 72.9& 70.8 & 70.1 & 70.94\\ 
 \hline
 \HTS   & 74.82  & 72.01  & 70.68 & 69.92  & 69.74 &   71.43 \\
 \hline
  \EthicalShort    & 68.41  & 60.4  & 60.8& 57.2  & 56.51 & 60.67 \\
  \hline
\end{tabular}
\caption {Identification: Indirect "gauge" of Recall: We report how many threads our method finds with 50\% keywords compared to the keyword based selection with larger sets of keywords [60-100]\% .} 
\label{tab:relevantRecall}
\end{table}

\begin{table}[t]
\centering
\small
\begin{tabular}{|l|c|c|c||c|} 
\hline
& \OffCommShort &  \HTS  &  \EthicalShort & Avg.\\ 
 \hline 
 Precision  & 98.2 & 97.5 & 97.0& 97.5\\ 
 \hline
\end{tabular}
\caption { Identification Precision: the precision of the identified thread of interest with the similarity-based method. }
\label{tab:relevantPercision}
\end{table}

{\bf Estimating precision.}
To evaluate precision,  we want to identify what percentage of the retrieved threads are relevant. To this end, we resort to manual evaluation. We have labeled 300 threads from each dataset retrieved with 50\% of the keywords and we get our annotators to identify if they are relevant. 
 We show the results in Table \ref{tab:relevantPercision}.
We understand that on average more than 97.5\% of the threads identified with the similarity based method are relevant
with an inter-annotator agreement 
Fleiss-Kappa coefficient of 0.952.

\begin{figure}[t]
  \includegraphics[width=1\linewidth]{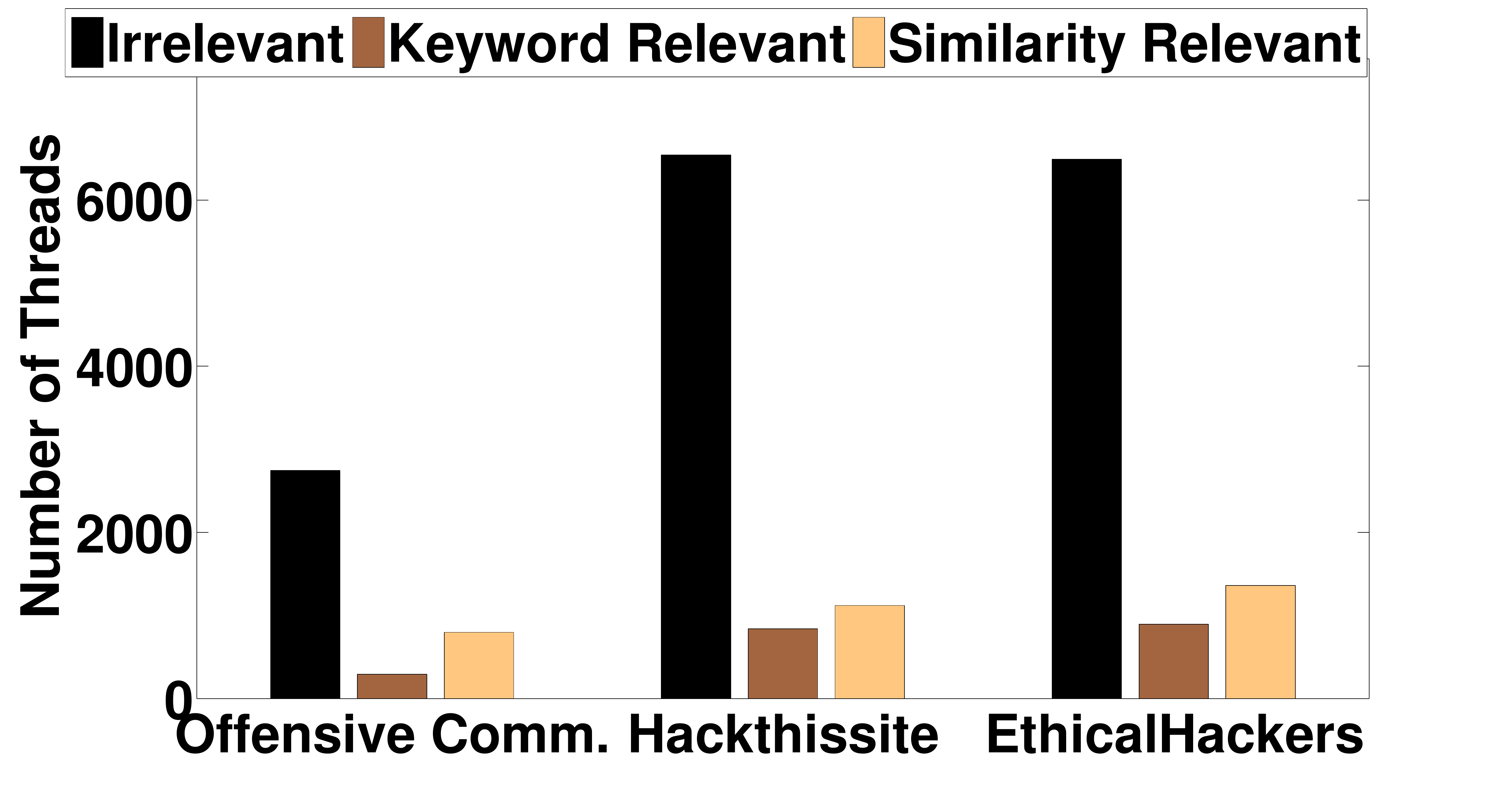}
  \centering
  \caption {Number of relevant thread in each forums identified by our approach: (a) irrelevant (not selected), (b) selected via keyword matching and (c) selected via similarity.}
    \label{fig:relevant}
\end{figure} 

{\bf The power of the embedding in determining similarity.}
We find that the similarity step identifies threads that are deemed relevant to a human reader, but are not  ``obviously similar", if you
examine the threads word for word.
We provide a few examples of threads that were identified by the keyword-based selection, and the related similar threads that our approach identified.
Table \ref{tab:SimilSample} and \ref{tab:SimilSample2} illustrate how the 
retrieved thread are similar to the target thread conceptually, without matching linguistically.    

{\bf  A four-fold  thread reduction.} Our approach reduces the amount of threads to only 22-25\% of the initial threads as shown in Table~\ref{tab:relevant}. 
Figure~\ref{fig:relevant} depicts the same data visually.
Clearly, these
results will vary depending on the keywords given by the user and the type of the forum.

\begin{table*}[ht]
 \centering 
 \small
 \begin{tabular}{|c|p{4cm}|p{8cm}|} 
 \hline
 Selection Method & Title & Post \\ 
 \hline
 Keyword selected &  [ULTIMATE] How to SPREAD your viruses successfully [TUTORIAL] & Educational Purposes NOT MINE  In this tutorial I will show you how to spread your trojans/viruses etc. I will show you many methods, and later you choose which one ....   \\ 
 \hline
 \multirow{3}{*}{Similarity selected } &  Botnet QA! & Just something I compiled quickly. Im also posting my bot setup guide soon. If you want any questions or links added to the Q\&A, please ask and Ill add them.   
\\\cline{2-3}
   &  The COMPLETE beginners guide to hacking &another great guide i found :D  Sections: 1) Introduction  2) The hacker manifesto  3) What is hacking?  4) Choosing your path  5) Where should I start?  6) Basic terminology  7) Keylogging... \\\cline{2-3}
   &  [TUT]DDoS Attack - A life lesson & Introduction I know their are a lot more ways to DoS than are shown here, but ill  let you figure them out yourself. If you find any mistake in this  tutorial please tell me^^ What is \'DDoS\'?   \\ 
 \hline
\end{tabular}
 \caption {Examples of similar threads for class \TT: threads offering hacking tutorials.}
 \label{tab:SimilSample}
\end{table*}

\begin{table*}[ht]
 \centering 
 \small
 \begin{tabular}{|c|p{4cm}|p{8cm}|} 
 \hline
 Selection Method & Title & Post \\ 
 \hline
 Keyword selected &  Blackmailed! How to hack twitter? & Hey, everyone. Im new on this website and I need help. Im trying to hack a twitter account because theyve been harassing me and reporting isnt helping at all.    \\ 
 \hline
 \multirow{4}{*}{Similariry selected} &  Need hacking services & IIm new to this website. Im not a hacker. Would like to hire hacking services to hack email account, Facebook account and if possible iPhone. Drop me a pm if you can do it. Fee is negotiable. Thanks   \\\cline{2-3}
   &  Hello hacker members &
    My name is XXXX and im looking for someone to help me crack a WordPress password from a site that has stolen all our copyrighted content. Weve reported to google but is taking forever.  I have the username of the site, just need help to crack the password so i can remove our content. Please message me with details if you can help   \\\cline{2-3}
   &  finding a person with his email & Hello guys! I need to find out how I can find a person \'behind\' an email! Let me explain please ...   \\\cline{2-3}
  & Hi & hello everyone im new here and i want to learn how to hack an account  any account in fact fb twitter even credit card hope you code help me out who knows maybe i can help you in the future right give and take \\
  \hline
\end{tabular}
 \caption {Examples of similar threads for class \PS: threads looking for hacking services.}
 \label{tab:SimilSample2}
\end{table*}


\subsection{Results 2: Thread Classification}

We present the results of our classification study.

{\bf \myalg compared to the state-of-the-art.} Our approach compares favourably against the competition. Table \ref{tab:results} summarizes the results of the baseline methods and our \myalg for three forums. \myalg outperforms other baseline methods with at least 1.4 percentage point in accuracy and 0.7 percentage point in F1 score, except BERT. First, using BERT ``right out of the box" did not give good results initially. However,  we fine-tuned BERT for this domain. 
BERT performs poorly on two sites, \HTS and \Ethical, while it performs well for \OffComm. We attribute this  to the limited training data in terms of text size and also the nature of the language users use in such forums. For example, we found that the titles of two  misclassified threads  contained typos and used unconventional slang and writing structure `` Hw 2 gt st4rtd with r3v3r53 3ngin33ring 4 n00bs!!'', ``metaXploit 3xplained fa b3ginners!!!''.  We intend to investigate BERT
 and how it can be tuned further  in future work. 
Note that methods BOW and NMF did not assign any instances to the minority classes correctly, therefore the value of F1 score in Table \ref{tab:results} is reported as NA.

\begin{table*}[ht]
 \centering 
\small
 \begin{tabular}{|l|r||r|r|r|r|r|r|| r|} 
 \hline
 Datasets & Metrics & BOW & NMF & SWEM & FastText & LEAM & BERT & {\bf REST} \\ 
 \hline
 \hline
 \multirow{2}{*}{\OffCommShort} & Accuracy  & 75.33$\pm$0.1  & 74.31$\pm$0.1  & 75.55$\pm$0.21 & 74.64$\pm$0.15 &  74.88$\pm$0.22 & \textbf{78.58$\pm$ 0.08} & 77.1$\pm$0.18 \\\cline{2-9}
   & F1 Score & NA & NA & 74.15$\pm$0.23 & 72.5$\pm$0.15 & 72.91$\pm$0.18 & \textbf{78.47$\pm$0.01}& 75.10$\pm$0.14\\ 
 \hline
 \hline
 \multirow{2}{*}{\HTS} & Accuracy & 65.3$\pm$0.41 & 69.46$\pm$0.12 &73.27$\pm$0.10 & 69.92$\pm$0.08 & 74.6$\pm$0.04 & 68.99$\pm$0.4 & \textbf{76.8$\pm$ 0.1} \\\cline{2-9}
   & F1 Score & NA & 70.23$\pm$0.13 & 71.89$\pm$0.14 & 65.81$\pm$0.4 & 71.41$\pm$0.09 &63.61$\pm$0.41& \textbf{74.47$\pm$0.24}\\
 \hline
 \hline
  \multirow{2}{*}{\Ethical} & Accuracy & 59.74$\pm$ 0.21 & 58.3$\pm$ 0.15 & 61.3$\pm$ 0.17 & 59.73$\pm$ 0.21 & 61.80 $\pm$0.13 & 54.91$\pm$ 0.32 & \textbf{63.3$\pm$  0.09} \\\cline{2-9}
   & F1 Score & NA & 57.83$\pm$0.16 & 59.6$\pm$0.23 & 59.5$\pm$0.13 & 60.9$\pm$0.17 & 51.78$\pm$0.15 & \textbf{61.7$\pm$0.21}\\
 \hline
\end{tabular}
 \caption { Classification: the performance of the five different methods in classifying threads in 10-fold cross validation. \label{tab:results}}
\end{table*}




\begin{figure}[htb!]
  \includegraphics[width=1\linewidth]{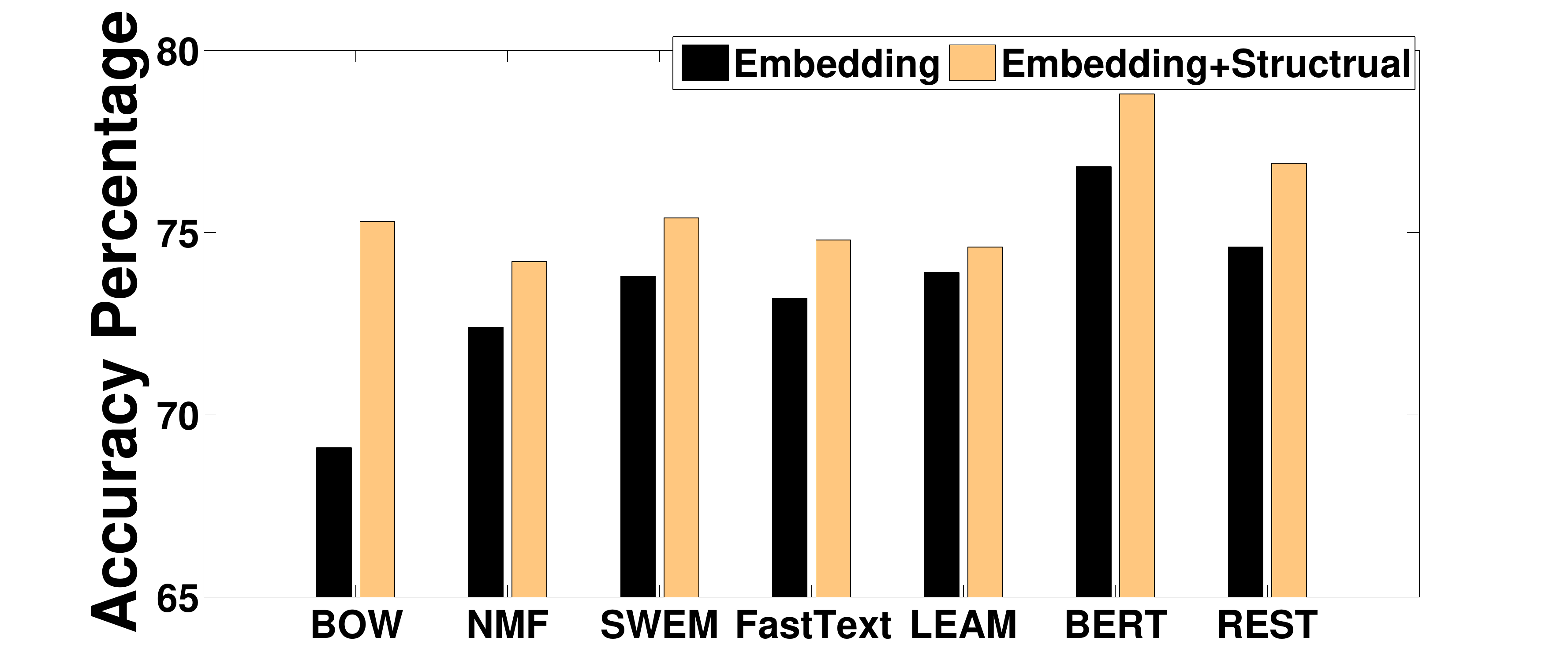}
  \centering
    \caption {Classification accuracy for two different features sets in 10-fold cross validation in \OffComm forum.}
  \label{fig:featureComp}
\end{figure}

{\bf The contextual features improves classification for all approaches.}
We briefly discussed contextual features in our classification section.
We conduced experiments with and without these features for all six algorithms and we show the results in Figure \ref{fig:featureComp}
for \OffComm.
Including the contextual features in our classification improves the
accuracy for all approaches (on average by 2.4\%). 
The greatest beneficiary is the Bag-of-Words method whose accuracy improves by roughly 6\%. 

\section{Related Work}
\label{sec:related}
We summarize related work group into areas of relevance.



{\bf a. Identifying entities of interest in security forums.}
Recently there have been a few efforts focused on extracting 
entities of interest in  security forums. 
A very interesting study focuses on the dynamics of the black-market of hacking goods and services and their pricing~\cite{Portnoff2017}, which for us is one of the categories of interest. Some other recent efforts focus on identifying malicious IP addresses that are reported in the forum~\cite{Joobin2018,Joobin2017}, which is relatively different task, as there, the entity of interest
has a well-defined format.
Another interesting work~\cite{Tavabi2018} uses a word embedding technique
focusing identifying vulnerabilities and exploits.

{\bf b. Identifying malicious users and  events.}
Several studies focus on identifying key actors and malicious users in security forums by utilizing their social and linguistics behavior. \cite{Li2014,Marin2018_keyhacker,abbasi2014}.
Another work~\cite{Gharibshah2019WWW,Sapienza2017_USC1} identifies emerging threats by 
monitoring threads activities and the behavior of malicious users and correlating it
with information from security experts on Twitter. Users' behaviors studied \cite{Zhabiz2020,eli2016} to identify abnormal users in interaction with each others.
Another study~\cite{Sapienza2018_USC2} detects emerging security concerns by monitoring the  keywords used in forums and other online platforms, such as blogs.

{\bf c. Analyzing other online forums.}
Researchers have analyzed a wide range of online forums such as blogs, commenting platforms, reddit etc. Indicatively,
we refer to a few recent studies. Google~\cite{Zhang2017} analyzed {\em question-answer} type of forums and they also published the large dataset
that they collected.
Another study  focusing on detecting question-answer threads within a discussion forum using linguistic features~\cite{Cong2008SIGIR}. 

Despite many common algorithmic approaches, we argue that each type of forum and different focus questions necessitate novel algorithms.

{\bf d. NLP, Bag-of-Words, and Word Embedding techniques.}
Natural Language Processing is a vast field, and even the more recent approaches, such as query transformation and word embedding have benefited from significant numbers of studies \cite{Scells2018,mccallum1998naive,Mikolov2013,Le2014_Doc2Vec,Li2014,Jin2016,Wang2018,Zamani2017,Shen2018,Lee1999}.
 ~Most recently, several methods focus on
 combining word embedding and deep learning approaches for text classification~\cite{Wang2018,Zamani2017,Shen2018,Bert2018}.

We now discuss the most relevant previous efforts. 
These efforts use word embedding representation and they use it for classification for text, but: (a) neither of those focuses on forums,
(b) there are some other technical differences with our work.
The first work, predictive text embedding (PTE) \cite{Tang2015}, 
uses a network-based approach, where each thread is described by a network
of interconnected entities (title, body, author etc).
The second study, LEAM \cite{Wang2018}, uses  a word embedding and a Neural Network classifier to create a thread embedding.
LEAM argues that it outperforms PTE, and as we show here, we outperform LEAM.
Recently Google introduced BERT~\cite{Bert2018},  a deep pre-trained bidirectional transformers for language understanding which uses a pre-trained unsupervised language model on large corpus of data. Although the power of large data set for training is indisputable, at the same time,
we saw first hand the need for some customization for each domain space.

Finally, there are some  efforts that use  Doc2Vec to identify the embedding of a document (equivalently threads in our case).
However, these techniques would not work well here due to the small size
of the datasets~\cite{Le2014_Doc2Vec}. This technique could be applied in much larger forums,
and we will consider it in such a scenario in the future.

\section{Conclusion}
There is a wealth of information in security forums, but still,
the analysis of security forums is in its infancy, despite several promising recent works.

 We propose a novel approach to identify and classify threads of interest
 based on a multi-step weighted word embedding approach. 
As we saw, our approach consists of two parts:
(a) a similarity-based approach  to extract  relevant threads reliably, and
b)  weighted embedding-based classification method to classify  threads of interest into user-defined classes.
The key novelty of the work is a multi-step  weighted embedding approach: we project words, threads and classes in the embedding space and establish relevance and similarity there. 

Our work is a first step towards developing an easy-to-use 
methodology that can harness some of the information in security forums.
The easy-of-use stems from the ability of our method to operate
with an initial set of bag-of-words, which our system uses
to seeds to identify threads that the user is interested in.
\section{ Acknowledgments}
This work is supported by  DHS ST Cyber Security (DDoSD)  HSHQDC-14-R-B00017 grant, NSF NeTS 1518878 and UC-NL-CRT LFR 18548554.



\bibliography{ref.bib}
\bibliographystyle{aaai}

\end{document}